\documentclass{sig-alternate-05-2015}
\usepackage{times}
\usepackage{helvet}
\usepackage{courier}
\usepackage{color}
\usepackage{graphicx}
\usepackage{caption}
\usepackage{subcaption}
\usepackage{float}
\usepackage{subfig}
\usepackage{array}
\usepackage{epsfig}
\usepackage{amsmath}
\usepackage{makecell}
\usepackage{algorithm}
\usepackage{algorithmic}
\usepackage{eqparbox}
\usepackage[utf8x]{inputenc}

\begin{document}

\CopyrightYear{2017} 
\setcopyright{acmlicensed}
\conferenceinfo{WSDM 2017,}{February 06 - 10, 2017, Cambridge, United Kingdom}
\isbn{978-1-4503-4675-7/17/02}\acmPrice{\$15.00}
\doi{http://dx.doi.org/10.1145/3018661.3018735}

\clubpenalty=10000 
\widowpenalty = 10000

\newcommand{\comments}[1]{\textcolor{red}{[\textit{#1}]}} 
\newcommand{\nop}[1]{}
\newcommand{\model}[0]{TP-HNE}
\title{Task-Guided and Path-Augmented Heterogeneous Network Embedding for Author Identification}

\numberofauthors{2}
\author{
\alignauthor Ting Chen$^{\thanks{Part of the work was done while author was at Northeastern University.}}$\\
       \affaddr{Department of Computer Science}\\
       \affaddr{University of California, Los Angeles}\\
       \email{tingchen@cs.ucla.edu}
\alignauthor Yizhou Sun\\
	   \affaddr{Department of Computer Science}\\
       \affaddr{University of California, Los Angeles}\\
       \email{yzsun@cs.ucla.edu}\\
}

\maketitle
\setlength{\parindent}{1em}

\begin{abstract}

In this paper, we study the problem of author identification under double-blind review setting, which is to identify potential authors given information of an anonymized paper. Different from existing approaches that rely heavily on feature engineering, we propose to use network embedding approach to address the problem, which can automatically represent nodes into lower dimensional feature vectors. However, there are two major limitations in recent studies on network embedding: (1) they are usually general-purpose embedding methods, which are independent of the specific tasks; and (2) most of these approaches can only deal with homogeneous networks, where the heterogeneity of the network is ignored. Hence, challenges faced here are two folds: (1) how to embed the network under the guidance of the author identification task, and (2) how to select the best type of information due to the heterogeneity of the network.

To address the challenges, we propose a task-guided and path-augmented heterogeneous network embedding model. In our model, nodes are first embedded as vectors in latent feature space. Embeddings are then shared and jointly trained according to task-specific and network-general objectives. We extend the existing unsupervised network embedding to incorporate meta paths in heterogeneous networks, and select paths according to the specific task. The guidance from author identification task for network embedding is provided both explicitly in joint training and implicitly during meta path selection. Our experiments demonstrate that by using path-augmented network embedding with task guidance, our model can obtain significantly better accuracy at identifying the true authors comparing to existing methods.
\end{abstract}


\vfill\eject  
\section{Introduction}

Heterogeneous networks are ubiquitous. Examples include bibliographic networks \cite{sun2011co,sun2012will}, movie recommendation networks \cite{yu2014personalized} and many online social networks containing information of heterogeneous types \cite{shi2015semantic}. Different from their homogeneous counterparts, heterogeneous networks contain multiple types of nodes and/or links. For example, in bibliographic networks, node types include paper, author and more; link types include author-write-paper, paper-contain-keyword and so on. Due to the fast emerging of such data, the problem of mining heterogeneous network has gained a lot of attention in the past few years \cite{sun2012mining,shi2015semantic}.

\begin{figure}[!t]
	\begin{center}
		\epsfig{file=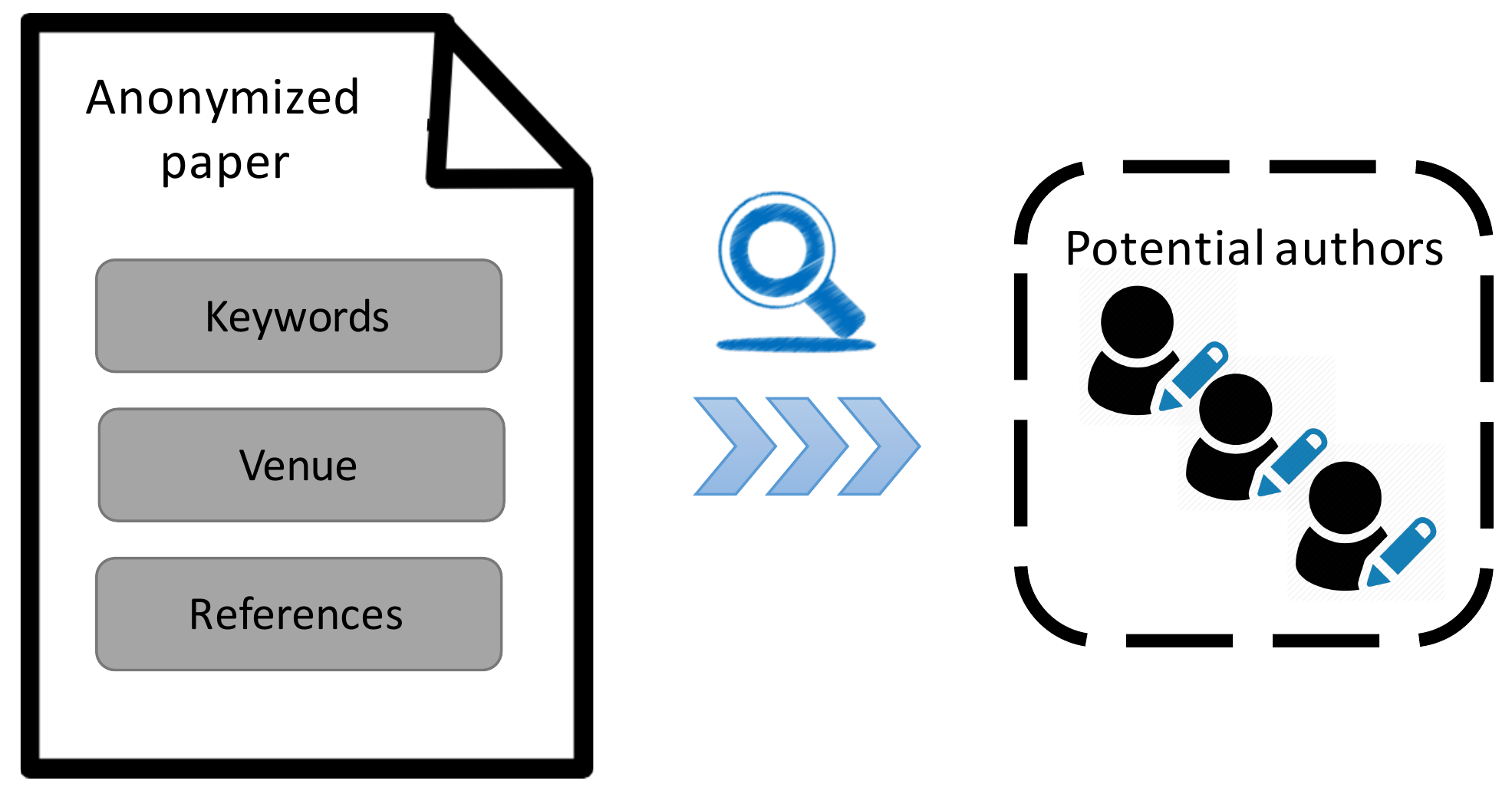,height=3.0cm}
	\end{center}
	\caption{\label{fig:toy} Illustration of author identification problem.}
\end{figure}

In this work, we are interested in the problem of mining heterogeneous bibliographic network \cite{sun2012mining}. To be more specific, we consider the problem of author identification under double-blind review setting \cite{hill2003myth}, on which many peer review conferences/journals are based. Authors of the paper under double-blind review are not visible to reviewers, i.e. the paper is anonymized, and only content/attributes of the paper (such as title, venue, text information, and references) are visible to reviewers. However, in some cases authors of the paper can still be unveiled by the content and references provided. Affected by the phenomenon, questions exist about whether or not double-blind review process is really effective. In fact, WSDM this year also conducts an experiment trying to answer this question. Here we ponder on this issue by formulating the author identification problem that aims at designing a model to automatically identify potential authors of an anonymized paper. Instead of dealing with full text directly, we treat the information of an anonymized paper as nodes in bibliographic network, such as keyword nodes, venue nodes, and reference nodes. An illustration of the problem can be found in Figure \ref{fig:toy}. Other than serving as a study for existing reviewing system, the problem has broader implications for general information retrieval and recommender system, where the model is asked to match queried document with certain target, such as reviewer recommendation \cite{van1999effect,sun2012will}.

To tackle the author identification problem, as well as many other network mining problems, good representations of data are very important, as demonstrated by many previous work \cite{mikolov2013distributed,mikolov2013efficient,perozzi2014deepwalk,tang2015line,chen2016entity}. Unlike traditional supervised learning, dense vectorized representations \cite{mikolov2013distributed,mikolov2013efficient} are not directly available in networked data \cite{tang2015line}. Hence, many traditional methods under network settings heavily rely on problem specific feature engineering \cite{li2013combination,li2013feature,zhong2013contextual,efimov2013kdd,zhao2013scorecard}.

Although feature engineering can incorporate prior knowledge of the problem and network structure, usually it is time-consuming, problem specific (thus not transferable), and the extracted features may be too simple for complicated data sets \cite{bengio2009learning}. Several network embedding methods \cite{perozzi2014deepwalk,tang2015line,tang2015pte} have been proposed to automatically learn feature representations for networked data. A key idea behind network embedding is learning to map nodes into vector space, such that the proximities among nodes can be preserved. Similar nodes (in terms of connectivity, or other properties) are expected to be placed near each other in the vector space.

Unfortunately, most existing embedding methods produce general-purpose embeddings that are independent of tasks, and they are usually designed for homogeneous networks \cite{perozzi2014deepwalk,tang2015line}. When it comes to author identification problem under the heterogeneous networks, existing embedding methods cannot be applied directly. There are two unique challenges brought by this problem: (1) how to embed the network under the guidance of author identification task, so that embeddings learned are more suitable for this task compared to general network embedding. And (2) how to select the best type of information due to the heterogeneity of the network. As shown in previous work \cite{sun2011pathsim,sun2012mining}, proximity in heterogeneous networks is richer than homogeneous counterparts, the semantic of a connection between two nodes is likely to be dependent on the type of connection they form.

To address the above mentioned challenges, we propose a task-guided and path-augmented network embedding method. In our model, nodes are first embedded as vectors. Then the embeddings are shared and jointly trained according both task-specific and network-general objectives: (1) the author identification task objective where embeddings are used in a specifically designed model to score possible authors for a given paper, and (2) the general heterogeneous network embedding objective where embeddings are used to predict neighbors of a node. By combing both objectives, the learned network can preserve network structures/proximities, as well as be beneficial to the author identification task. To better utilize the heterogeneous network structure, we extend the existing unsupervised network embedding to incorporate meta paths derived from heterogeneous networks, and select useful paths according to the author identification task. Compared to traditional network embedding \cite{perozzi2014deepwalk,tang2015line,tang2015pte}, our method uses the author identification task as an explicit guidance to influence network embedding by joint learning, and also as an implicit guidance to select meta paths, based on which the network embedding is performed. It is worth mentioning that although our model is originally targeted for the author identification problem, it can also be extended to other task-oriented embedding problems in heterogeneous networks.

The contributions of our work can be summarized as follows.

\begin{itemize}
\item We propose a task-guided and path-augmented heterogeneous network embedding framework, which can be applied to author identification problem under double-blind review setting and many other tasks.
\item We demonstrate the effectiveness of task guidance for network embedding when a specific task is of interest; and also show the usefulness of meta-path selection in heterogeneous network embedding.
\item Our learning algorithm is efficient, parallelizable, and experimental results show that our model can achieve much better results than existing feature based methods.
\end{itemize}
\section{Preliminaries}
\label{sec:problem}

In this section, we first introduce the concept of heterogeneous networks and meta paths, and then introduce the embedding representation of nodes. Finally, a formal definition of the author identification problem is given.

\subsection{Heterogeneous Networks}

\textbf{Definition 1 (Heterogeneous Networks)} A \textit{heterogeneous network} \cite{sun2012mining} is defined as a network with multiple types of nodes and/or multiple types of links. It can be denoted as $G=(\mathcal{V}, \mathcal{E})$, where $\mathcal{V}$ is a set of nodes and $\mathcal{E}$ is a set of links. A heterogeneous network is also associated with a node type mapping function $f_v: \mathcal{V} \rightarrow \mathcal{O}$, which maps the node to a predefined node type, and a link type mapping function $f_e: \mathcal{E} \rightarrow \mathcal{R}$, which maps the link to a predefined link type. It is worthing noting that a link type automatically defines the node types of its two ends.

\begin{figure}[h]
\begin{center}
\epsfig{file=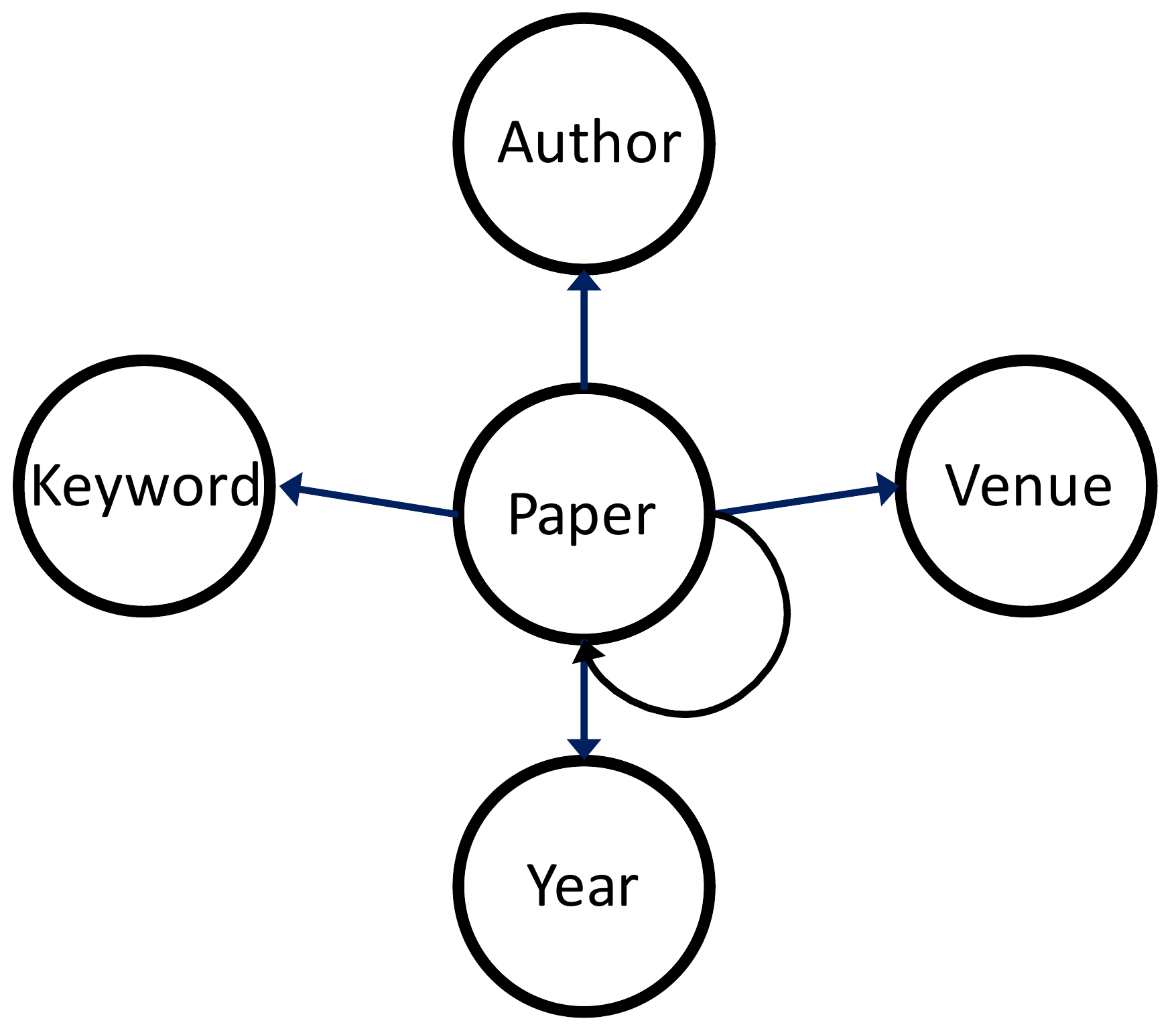,height=4.0cm}
\end{center}
\caption{\label{fig:net_schema}Network schema of the heterogeneous bibliographic network. Each node denotes a node type, and each link denotes a link type.}
\end{figure}

The bibliographic network can be seen as a heterogeneous network \cite{sun2012mining}. It is centered by paper, the information of a paper can be represented as its neighboring nodes. The node types $\mathcal{N}$ in the network include paper, author, keyword, venue and year, and the set of link types $\mathcal{R}$ include author-write-paper, paper-contain-keyword, and so on. The network schema is shown in Figure \ref{fig:net_schema}.

\textbf{Definition 2 (Meta path)} A \textit{meta path} \cite{sun2011pathsim} is a path defined on the network schema $T_G = (\mathcal{O}, \mathcal{L})$ and is denoted in the form of $o_1 \xrightarrow{l_1} o_2 \xrightarrow{l_2} \cdots \xrightarrow{l_m} o_{m+1}$, which represents a compositional relations between two given types. For each of the meta path $r$, we can define an adjacency matrix $M^{(r)}$, with cardinality equal to the number of nodes, to denote the connectivity of nodes under that meta path. If there are multiple meta paths considered for a given network $G$, we use a set of adjacency matrices $\{M^{(r)}\}$ to represent it.

Examples of meta paths defined in network schema Figure \ref{fig:net_schema} include paper $\rightarrow$ keyword $\leftarrow$ paper, and paper $\rightarrow$ year $\leftarrow$ paper. From these two examples, it is easy to see that in a heterogeneous network, even compare two nodes of the same type (e.g. paper), going from different paths can lead to different semantic meanings.

\subsection{Embedding Representation of Nodes}

The networked data is usually high-dimensional and sparse, as there can be many nodes but the links are usually sparse \cite{albert2002statistical}. This brings challenges to represent nodes in the network. For example, given two users, it is hard to calculate their similarity or distance directly. To obtain a better data representation, embedding methods are widely adopted \cite{perozzi2014deepwalk,tang2015line,tang2015pte}, where nodes in the network are mapped into some common latent feature space. With embedding, we can measure similarity/distance between two nodes directly based on arithmetic operations, like dot product, of their embedding vectors. 

Through the paper, we use a matrix $U$ to represent the embedding table for nodes. The size of the matrix is $N \times D$, where $N$ is total number of nodes (including all node types, such as authors, keywords, and so on), and $D$ is the number of dimensions. So the feature vector for node $n$ is denoted as $u_n$, which is a $D$-dimensional vector.

\subsection{Author Identification Problem}

We formalize the author identification problem using bibliographic networks with network schema shown in Figure \ref{fig:net_schema}. For each paper $p$, we represent its neighbors in the given network $G$ as $X_p = \{X_p^{(1)}, X_p^{(2)}, \cdots, X_p^{(T)}\}$, where $X_p^{(t)}$ is a set of neighbor nodes in $t$-th node type. The node types include keyword, reference, venue, and year in our task. And we use $A_p$ to denote the set of true authors of the paper $p$.

\textit{\textbf{Author Identification Problem}}. Given a set of papers represented as $(X, A)$ where $X = \{X_p\}, A = \{A_p\}$, the goal is to learn a model to rank potential authors for every anonymized paper $p$ based on information in $X_p$, such that its top ranked authors are in $A_p$\footnote{Here it is posed as a ranking problem since each paper may have different number of authors and it is unknown beforehand.}.

\section{Proposed Model}

In this section, we introduce the proposed model in details. The model is composed of two major components: (1) author identification based on task-specific embedding, and (2) path-augmented general network embedding. We first introduce them separately and then combine them into a single unified framework, where the meta paths are selected according to the author identification task.

\subsection{Task-Specific Embedding for Author Identification}
\label{sec:sup_only}

In this subsection, we propose a supervised embedding-based model that can rank the potential authors given the information of a paper (such as keywords, references, and venue). Our model first maps each node into latent feature space, and then gradually builds the feature representation for the anonymized paper based on its observed neighbors in the network. Finally the aggregated paper representation is used to score the potential author.

There are two stages of aggregation to build up the feature representation for a paper $p$ based on node embeddings. In the first stage, it builds a feature vector for each of the $t$-th node type by averaging node embeddings in $X^{(t)}_p$, which is:
\begin{equation}
V^{(t)}_p = \sum_{n \in X^{(t)}_p} u_n / |X^{(t)}_p|
\end{equation}
where $V^{(t)}_p$ is the feature representation of $t$-th node type (e.g. keyword node type), and $u_n$ is the $n$-th node embedding (e.g. keyword node).

In the second stage, it builds feature vector for the paper $p$ using a weighted combination of feature vectors of different node types:
\begin{equation}
V_p = \sum_{t} w_t V^{(t)}_p
\end{equation}

Now the anonymized paper $p$ is represented by this feature vector $V_p$, and can be used to score potential authors (which are also embedding vectors) by taking their dot product. The score between a pair of paper and author is defined as follows:
\begin{equation}
\begin{split}
f(p,a) &= u^T_a V_p = u^T_a \bigg( \sum_t w_t V^{(t)}_p \bigg) \\
&= u^T_a \bigg( \sum_t w_t \sum_{n \in X^{(t)}_p} u_n / |X^{(t)}_p| \bigg)
\end{split}
\end{equation}

\begin{figure}[!t]
\begin{center}
\epsfig{file=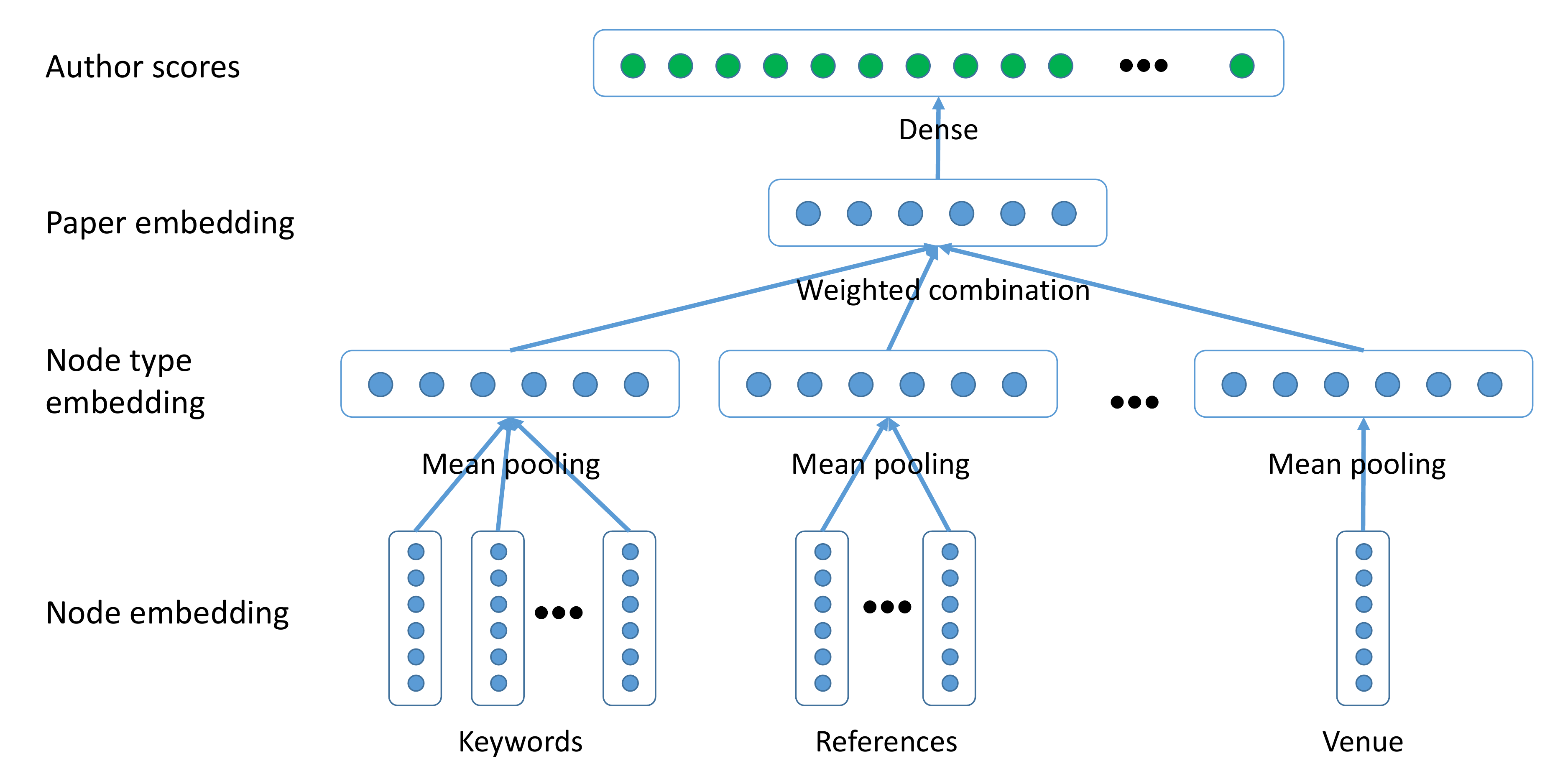,height=4.2cm}
\end{center}
\caption{\label{fig:sup_net} Task-specific embedding architecture for author identification.}
\end{figure}

The computational flow is summarized in Figure \ref{fig:sup_net}. Note that the final densely-connected layer has no bias term, and thus its weight matrix can be seen as author node embeddings. The final layer output (green dots) is the score vector for the candidate authors.

To learn the parameters $U$ and $w$, we use stochastic gradient descent (SGD) \cite{bottou2010large} based on a hinge loss ranking objective. For each triple $(p, a, a')$, where $a$ is one of the true author for paper $p$, and $a'$ is not the author of paper $p$, the hinge loss is defined as:
\begin{equation}
\label{eq:hinge_single}
\text{max}\bigg(0, f(p,a') - f(p,a) + \xi\bigg)
\end{equation}
where $\xi$ is a positive number usually referred as margin \cite{bordes2013translating}. A loss penalty will incur if the score of positive pair $f(p, a)$ is not at least $\xi$ larger than the score of $f(p, a')$.

To sample a triple $(p, a, a')$ used in SGD, we randomly select a paper $p$ from $X_p$ and one of its author $a$ from $A_p$, then sample a negative author from the pre-defined noise distribution $a' \sim P^{author}_n(a')$, such as discrete distribution based on author degree (with a similar idea of unigram distribution in word2vec \cite{mikolov2013distributed,mikolov2013efficient}). 

\subsection{Path-Augmented General Heterogeneous Network Embedding}
\label{sec:net_only}

In this subsection, we propose a path-augmented general network embedding model to exploit the rich information in heterogeneous networks.

Most of existing network embedding techniques \cite{perozzi2014deepwalk,tang2015line,tang2015pte} are based on the idea that, embeddings of nodes can be learned by neighbor prediction, which is to predict the neighborhood given a node, i.e. the linking probability $P(j|i)$ from node $i$ to node $j$. For existing network embedding methods, the observed neighborhood of a node is usually defined by original network \cite{tang2015line,tang2015pte} or by random walk on the original network \cite{perozzi2014deepwalk}.

In heterogeneous network, one can easily enrich the semantic of neighbors by considering different types of meta paths \cite{sun2011pathsim}. As shown in \cite{sun2011pathsim}, different meta paths encode different semantic of links. For example, connections between two authors can encode multiple similarities: (1) they are interested in the same topic, or (2) they are associated with the same affiliation. And clearly these two types of connections indicate different semantics. Inspired by the phenomenon, we generalize existing network embedding techniques \cite{tang2015pte} to incorporate different meta paths, and propose the path-augmented network embedding.

In path augmented network embedding, instead of using original adjacency matrices $\{E^{(l)}\}$ where $l$ is an original link type or one-hop meta path (such as author$\rightarrow$write$\rightarrow$paper), we consider more meta paths (such as author$\rightarrow$write$\rightarrow$paper$\rightarrow$contain$\rightarrow$keyword) and use meta path-augmented adjacency matrices $\{M^{(r)}\}$ for network embedding, where each $M^{(r)}$ indicates network connectivity under a specific meta path $r$. Here we normalize each $M^{(r)}$, such that $\forall r, \sum_{i,j} M^{(r)}_{i,j} = 1$, so that the learned embedding will not be dominated by some meta paths with large raw weights. Since there can be infinite many potential meta paths (including original link types), when considered for network embedding, one has to select a limited number of useful meta paths. The selection of meta paths will be discussed in next sub-section, and we assume a collection of meta paths are selected for now.

To learn embeddings that preserve proximities among nodes induced by meta paths, we follow the neighbor prediction framework, and model the conditional neighbor distribution of nodes. In heterogeneous networks, there can be multiple types of paths starting from a node $i$, so the neighbor distribution of the node will be conditioned on both the node $i$ and the given path type $r$, which is defined as follows:
\begin{equation}
\label{eq:unsup_condition}
P(j|i;r) = \frac{\exp(u_{i}^T u_{j})}{\sum_{j'\in DST(r)} \exp(u_{i}^T u_{j'})}
\end{equation}
where $u_i$ is the embedding of node $i$, and $DST(r)$ denotes the set of all possible nodes that are in the destination side of path $r$.

In real networks, the number of nodes in $DST(r)$ can be very large (e.g. millions of papers), so the evaluation of Eq. \ref{eq:unsup_condition} can be prohibitively expensive. Inspired by \cite{mikolov2013distributed,mikolov2013efficient}, we apply negative sampling and form the following approximation term:
\begin{equation}
\begin{split}
\label{eq:unsup_condition_app}
\log\hat{P}(j|i;r) \approx& \log \sigma(u_{i}^T u_{j}+b_r) +\\
& \sum_{l=1}^{k} \mathbb{E}_{j' \sim P^r_n(j')}[\log \sigma(-u_{i}^T u_{j'}-b_r)]
\end{split}
\end{equation}
where $j'$ is the negative node sampled from a pre-defined noise distribution $P^r_n(j')$ for path $r$ \footnote{The noise distribution only returns nodes of the same type as specified by end-point of path $r$.}, and a total of $k$ negative nodes are sampled for each positive node $i$. Furthermore, a bias term $b_r$ is added to adjust densities of different paths.

To learn the parameters $U$ and $b$, we adopt stochastic gradient descent (SGD) with the goal of maximizing the likelihood function. The training procedure is given as follows. We first sample a path $r$ uniformly, and then randomly sample a link $(i, j)$ according to their weights in $M^{(r)}$. The set of negative nodes $\{j'\}$ used in Eq. \ref{eq:unsup_condition_app} are also sampled according to some pre-defined $P^r_n(j')$, such as ``smoothed" node degree distribution under specific edge type \cite{mikolov2013distributed,mikolov2013efficient}. Finally the parameters $U, b$ are updated according to their gradients, such that approximated sample log-likelihood $\log \hat{P}(j|i;r)$ can be maximized. 

\subsection{The Combined Model}

The task-specific embedding sub-model and path-augmented general embedding sub-model capture different perspectives of a network. The former focuses more on the direct information related to the specific task, while the latter can better explore more global and diverse information in the heterogeneous information network. This motivates us to model them in a single unified framework.

The two sub-models are combined in two levels as follows.

\begin{itemize}
\item A joint objective is formed by combining both task-specific and network-general objectives, and joint learning is performed. Here the task serves as an explicit guidance for network embedding.
\item The meta paths used in network-general embedding are selected according to the author identification task. Here the task provides an implicit guidance for network embedding as it helps select meta paths.
\end{itemize}

\subsubsection{Joint Objective - An Explicit Guidance}

The joint objective function is defined as a weighted linear combination of the two sub-models with a regularization term on the embedding, where the embedding vectors are shared in both sub-models:
\begin{equation}
\label{eq:joint}
\begin{split}
\mathcal{L} =&  (1-\omega) \mathcal{L}_{task-specific} + \omega \mathcal{L}_{network-general} + \Omega(\mathcal{M}) \\
=& (1-\omega) \mathbb{E}_{(p, a, a')}\bigg[\text{max}\bigg(0, f(p,a') - f(p,a) + \xi\bigg)\bigg] \\
&+ \omega \mathbb{E}_{(r, i, j)}\bigg[-\log \hat{P}(j|i;r)\bigg] + \lambda \sum_i \|u_i\|^2_2
\end{split}
\end{equation}
where $\omega \in [0, 1]$ is the trade-off factor for task-specific and network-general components. When $w=1$, only network-general embedding is used; and when $w=0$, only supervised embedding is used. A regularization term is added to avoid over-fitting.

To optimize the objective in Eq. \ref{eq:joint}, we utilize Asynchronous Stochastic Gradient Descent (ASGD), where samples are randomly drawn and training is performed in parallel \cite{mikolov2013distributed}. The challenge here is that we have two different tasks that learn from two different data sources. To solve this problem, we design a sampling based task scheduler. Basically, for each worker, it first draws a task according to $\omega$, and then draws samples for the selected task and update the parameters according to the samples. In order to reduce the task sampling overhead, the selected task will be trained on a mini-batch of data samples instead of on a single sample.

The learning algorithm is summarized in Algorithm \ref{alg:framework}.
\begin{algorithm}[h]
\caption{Learning Framework}
\label{alg:framework}
\renewcommand{\algorithmicrequire}{\textbf{Input:}}
\renewcommand\algorithmicensure {\textbf{Output:}}
\begin{algorithmic}[1]
\REQUIRE Training data $X, A$ and path-augmented adjacency matrices $\{M^{(r)}\}$.
\ENSURE Parameters $U, w, b$
\WHILE{not converged}
\FOR{each thread}
	\STATE Sample one of the two tasks $\sim Bern(\omega)$
	\IF{the taks is network-general embedding}
		\STATE sample a mini-batch of $(r, i, j)$ triplets
		\STATE sample negative nodes $\{j'\}$
		\STATE update parameters $U, w$ according to their gradients
	\ELSE[the task is author identification]
		\STATE sample a mini-batch of $(p, a, a')$ triplets
		\STATE update parameters $U, b$ according to their gradients
	\ENDIF
\ENDFOR
\ENDWHILE
\end{algorithmic}
\end{algorithm}

\textbf{Complexity}. Firstly, the algorithm can be run in parallel using multiple CPUs thanks to asynchronous SGD. Secondly, the algorithm is efficient, as for each iteration of each thread, there are two major components: (1) both edge and negative node sampling only take constant time with alias table \cite{walker1977efficient}, and (2) gradient update is linear w.r.t. the number of links and number of embedding dimensions. Thirdly, with mini-batch of reasonable size, the overhead in switching tasks is ignorable.

\subsubsection{Meta Path Selection - An Implicit Guidance}
\label{sec:path_select}

So far we have assumed that path-augmented adjacency matrices $\{M^{(r)}\}$ are already provided. Now we discuss how we can select a set of meta paths that can further enhance the performance of the author prediction task.

The potential meta paths induced from the heterogeneous network $G$ can be infinite, but not every one is relevant and useful for the specific task of interest. So we utilize the author identification task as a guidance to help select the meta paths that can best help the task at hand. 

The path selection problem can be formulated as: given a set of pre-defined candidate paths $R=\{r_1, r_2, \cdots, r_L\}$, we want to select a subset of paths $R_{selected} \subseteq R$, such that certain utility can be maximized. Since our final goal is the author identification task, we define the utility to be the generalization performance (on validation data set) of the task.

It is worth noting the problem is neither differentiable nor continuous. And the total number of combinations are exponential to the number of candidate paths. So we employ following two steps to select relevant paths in a greedy fashion.

\begin{itemize}
\item[1] Single path performance. We first run the joint learning with network embedding based on a single path at a time, and then run the experiments for all candidate paths.
\item[2] Greedy additive path selection. We sort paths according to their performance (from good to poor) obtained from Step 1 above, and gradually add paths into the selected pool. Experiments are run for each additive combination of paths, and the path combination with best performance is selected.
\end{itemize}

We need to run experiments (at most) $2N$ times, where $N$ means the number of candidate paths. Since every experiment takes about 10 minutes in our case (even with millions of nodes and hundreds of millions of links), such selection scheme is affordable and can avoid exponential number of combinations.
\section{Experiments}

In this section, we compare the proposed model with baselines, and also evaluate several variants of the proposed model. Case studies are also provided.

\subsection{Data}

The AMiner citation network \cite{tang2008arnetminer} is used throughout our experiments. To prepare for the evaluation, we split all papers into training set and test set according to their publication time. Papers published before 2014 are treated as training set, and papers published in 2014 and 2015 are treated as test set. 

Based on the training papers, a heterogeneous bibliographic network is extracted. We first extract all papers which contain information about its title, authors, references, venue from the dataset. Then we extract keywords by combining unigram and key phrases extracted using method proposed in \cite{sigmod15_liu}. The schema of the network is the same as in Figure \ref{fig:net_schema}.

The extracted network contains millions of nodes and tens of millions of links. The detailed statistics of nodes and links for both training and test set can be found in Table \ref{tab:nodes} and \ref{tab:l1_path}, respectively.

\begin{table}[t]
\small
\centering
\caption{Node statistics}
\label{tab:nodes}
\begin{tabular}{|c|c|c|c|c|c|}
\hline
& Paper   & Author  & keyword   & Venue & Year \\ \hline \hline
Train &  1,562,139 &  1,003,836 &   402,687 &   7,528 &    60 \\\hline
Test  &    33,644 &    62,030 &    41,626 &    868 &     2 \\ \hline
\end{tabular}
\end{table}

\begin{table}[t]
\small
\centering
\caption{Length-1 link statistics}
\label{tab:l1_path}
\begin{tabular}{|c|c|c|c|c|c|}
\hline
&P-A & P-P & P-V & P-W & P-Y \\ \hline\hline
Train & 4,554,740 & 6,122,252 & 1,562,139 & 12,817,479 & 1,562,139 \\ \hline
Test & 96,434 & 388,030 & 235,508 & 287,885 & 235,508 \\ \hline
\end{tabular}
\end{table}

\textbf{Meta paths augmentation}. Other than the length-1 paths presented in the original network, we also consider various of length-2 meta paths as candidate paths for general heterogeneous network embedding. Although other path similarity measures \cite{sun2011pathsim} can be explored, for simplicity, we set weights of a path by the number of path instances. For example, if Tom attended KDD Twice and Jack attended KDD three times, then the path of Tom - KDD - Jack will have a weight of six. The augmented network by adding new meta paths has hundreds of millions of links, much more than the original network. Many of the candidate paths are not symmetric and may contain different information at both sides, so we consider them in both directions. Finally, the detailed statistics of the length-2 paths are presented in Table \ref{tab:l2_path}.

\begin{figure*}[]
	\begin{center}
		\epsfig{file=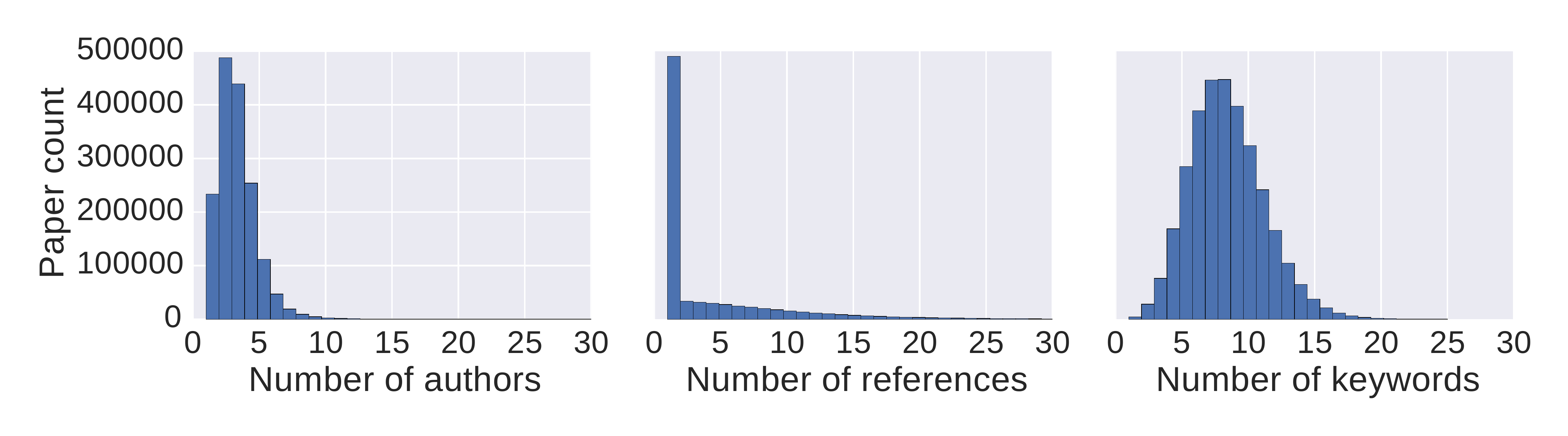,width=.7\textwidth}
	\end{center}
	\caption{\label{fig:net_stat} Distributions of numbers of authors, references and keywords.}
\end{figure*}

\begin{table*}[]
\small
\centering
\caption{Length-2 link statistics}
\label{tab:l2_path}
\begin{tabular}{|c|c|c|c|c|l|l|l|l|l|}
\hline
A-P-A & A-P-P & A-P-V & A-P-W & A-P-Y & P-P-V & P-P-W & V-P-W & W-P-W & Y-P-W \\ \hline \hline
17,205,758 & 18,308,110 & 4,554,740 & 38,251,803 & 4,554,740 & 3,674,632 & 27,200,144 & 12,817,479 & 118,497,737 & 12,817,479 \\ \hline
\end{tabular}
\end{table*}

To better understand statistics of the network,, Figure \ref{fig:net_stat} shows three different types of degree distributions for papers. As can be seen from the figure, most papers contain quite sparse information of authors, references and keywords: medium 3 authors, 1 reference (many are missing in the data set), and 8 keywords. And this lack of information makes the problem of automatic author identification even harder.

\subsection{Baselines and Experimental Settings}

We mainly consider two types of baselines: (1) the traditional feature-based methods, and (2) the variations of network embedding methods.

\begin{itemize}
\item \textbf{Supervised feature-based baselines}. As widely used in similar author identification/disambiguation problems \cite{li2013combination,li2013feature,zhong2013contextual,efimov2013kdd,zhao2013scorecard}, this thread of methods first extract features for each pair of training data, and then applies supervised learning algorithm to learn some ranking/classification functions. Following them, we extract 20+ related features for each pair of paper and author in the training set (details can be found in appendix). Since the original network only contains true paper-author pairs, in order to get the negative samples, for each paper-author pair we sampled 10 negative pairs by randomly replacing the authors. For the supervised algorithm, we consider Logistic Regression (\textbf{LR}), Support Vector Machines (\textbf{SVM}), Random Forests (\textbf{RF}), and \textbf{LambdaMART} \footnote{for LR, SVM, RF, we use scikit learn implementation, and for LambdaMART, we use XGboost implementation.}. For all these methods, we use grid search to find their best hyper-parameters, such as regularization penalty, maximum depth of trees, and so on.
\item \textbf{Task-specific embedding}. This method is introduced in Section \ref{sec:sup_only}. The embeddings of nodes are learned solely based on task-specific embedding architecture.
\item \textbf{Network-general embedding}. This method is introduced in Section \ref{sec:net_only}. The embeddings of nodes are learned solely based on general heterogeneous network embedding, and then the learned embeddings are used to score the author in the same way as in task-specific author identification framework. Since it is not directly combined with author identification task, it cannot perform path selection specific for the task. By default, the paths used for embedding are from original network, i.e. length-1 paths. With length-1 paths, this method is in the same form of PTE \cite{tang2015pte}.
\item \textbf{Pre-training + Task-specific embedding}. Pre-training has been found useful to improve neural network based supervised learning \cite{erhan2010does}. So instead of training task-specific author identification from randomly initialized embedding vectors, we first pre-train the embedding of nodes using network-general embedding, and then initialize the supervised embedding training with the pre-trained embedding vectors.
\item \textbf{Proposed combined model}. This is our proposed method, which combines both task-specific embedding and meta-path selection-based network-general embedding.
\end{itemize}

\textbf{Candidate authors.} There are more than one million authors in the training data, so the total number of candidate authors for each paper is very large. The supervised feature-based baselines cannot scale up to such large amount of candidate set, as it is both very time consuming and storage intensive to extract and store features for all candidate paper-author pairs (which amounts to more than $10^{17}$ pairs). Hence, we conduct comparisons mainly based on a sub-sampled author candidate set, where we randomly sample a set of negative authors, combined with the true authors of the paper to form a candidate set of total 100 authors. For completeness, we also provide both quantitative and qualitative comparisons of different embedding methods on the whole candidate set of a million authors.

\subsection{Evaluation Metrics}

Since the author identification problem is posed as a ranking problem and usually only top returned results are of interest, we adopt two commonly used ranking metrics: Mean Average Precision at k (MAP@k) and Recall at k (Recall@k).

MAP@K reflects the accuracy of top ranked authors by a model, and can be computed as mean of AP@K for each papers in the test set. The formula for computing AP@K of a single paper is given as follows.
\begin{equation}
\begin{split}
AP@K = \sum_{k=1}^{K} P(k) / \text{min}(L, K)
\end{split}
\end{equation}
where $P(k)$ is the precision at cut-off $k$ in the return list. $L$ is the total number of true authors for this test paper.

The Recall@K shows the ratio of true authors being retrieved in the top k return results, and can be computed according to:
\begin{equation}
\begin{split}
Recall@K = \frac{\text{\# of true authors at top K}}{\text{\# of total true authors}}
\end{split}
\end{equation}

\subsection{Meta-Path Selection Results}

We first report experimental results for path selection since the selected paths are used in the joint training of our model. The candidate paths that we consider are all length-1 and length-2 paths presented in Table \ref{tab:l1_path} and \ref{tab:l2_path}, 15 paths in total. As introduced in section \ref{sec:path_select}, a greedy algorithm involving two stages has been used for path selection: (1) single path performance evaluation, and (2) additive path selection.

\begin{figure}[t]
        \centering
        \begin{subfigure}[b]{0.4\textwidth}
                \includegraphics[width=\textwidth]{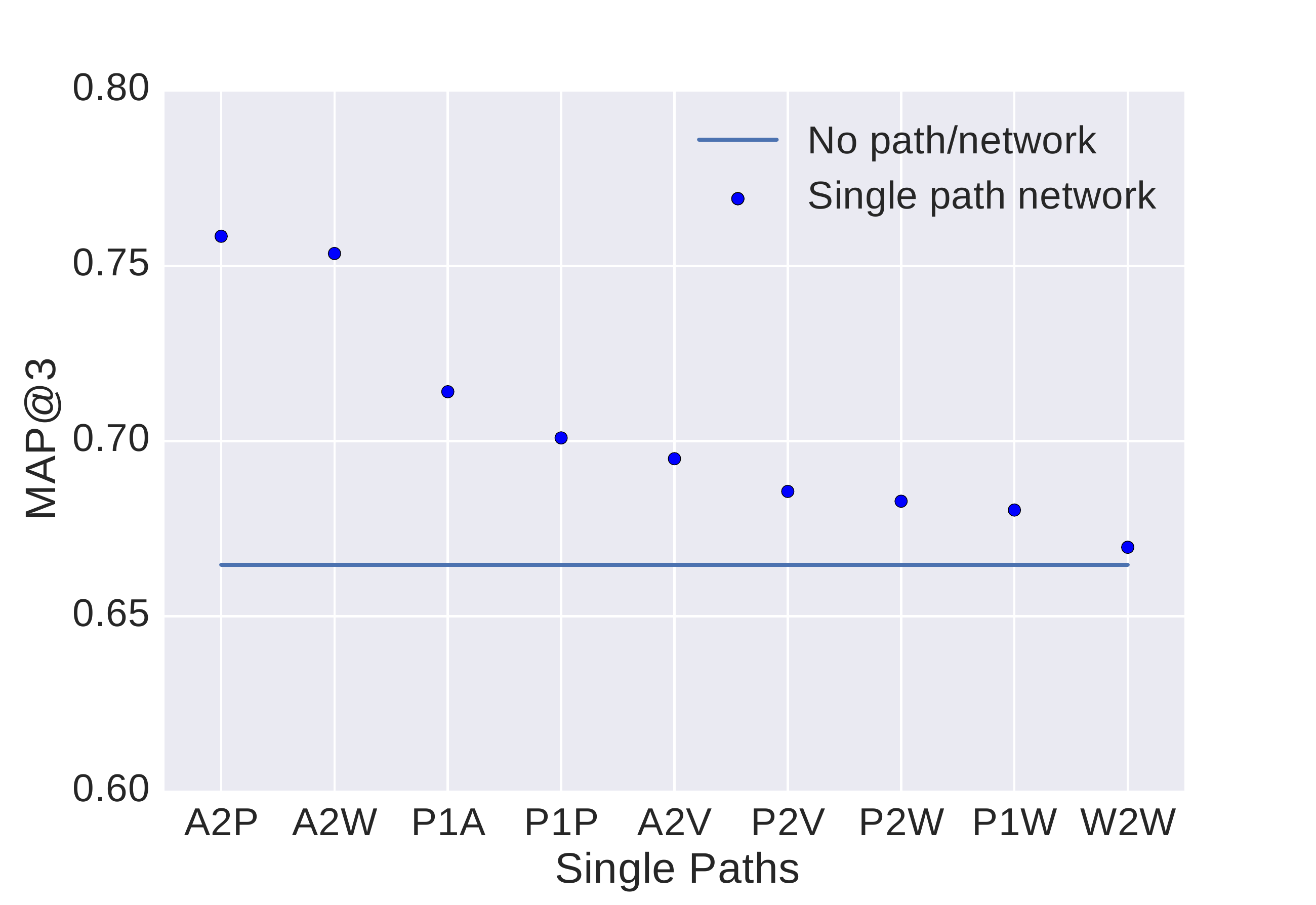}
                \caption{Single path performance.}
                \label{fig:path_select_a}
        \end{subfigure}
        \begin{subfigure}[b]{0.4\textwidth}
                \includegraphics[width=\textwidth]{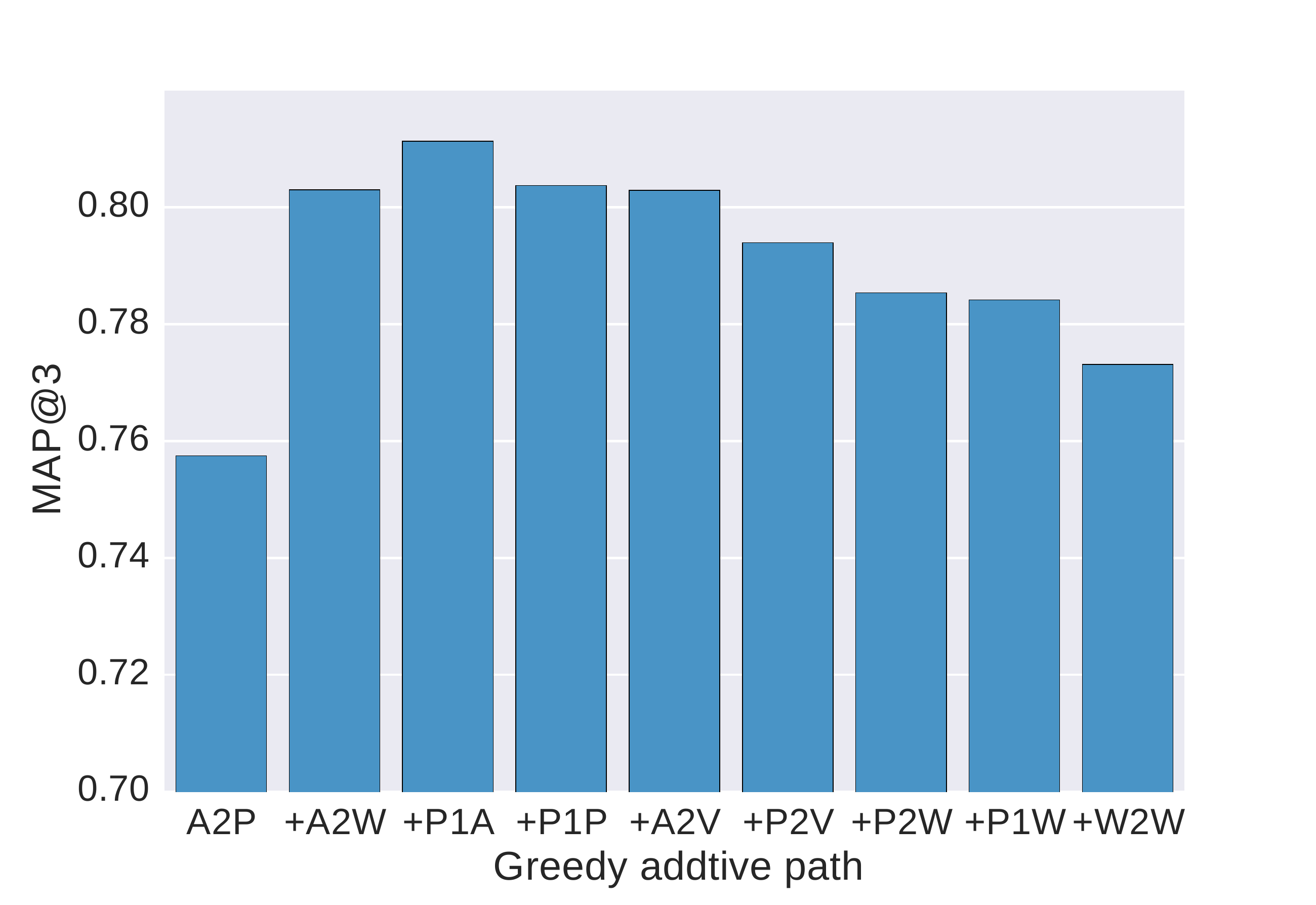}
                \caption{Additive selection performance. }
                \label{fig:path_select_b}
        \end{subfigure}
        \caption{Path selection under task guidance. Path names are shorten. X2Y denotes length-2 path; X1Y denotes length-1 path.}\label{fig:path_select}
\end{figure}

\begin{figure*}
	\centering
	\makebox[1.35\linewidth][c]{%
		\hspace{-19em}
		\begin{subfigure}[b]{0.26\textwidth}
			\includegraphics[width=\textwidth]{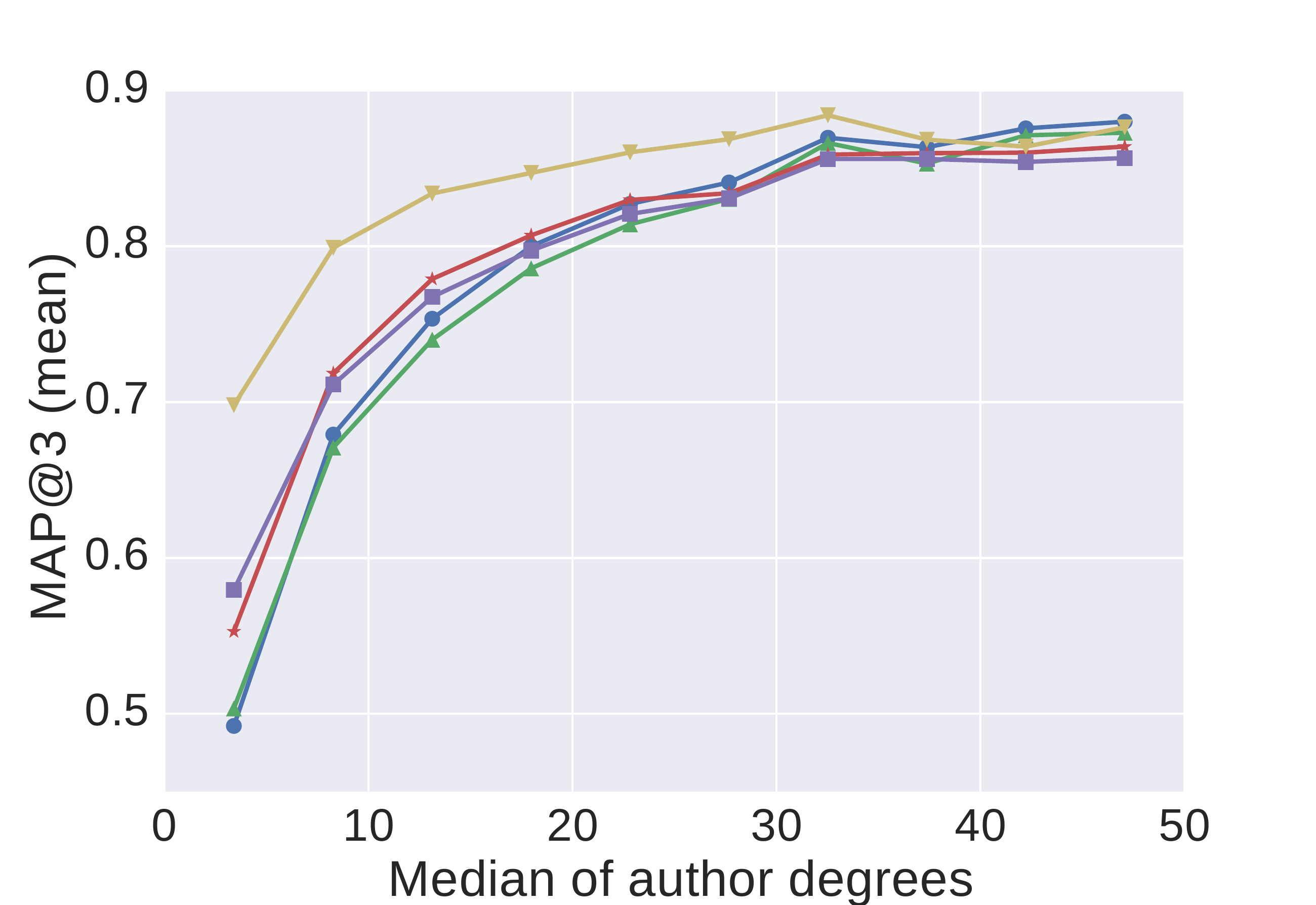}
			\caption{MAP@3}
		\end{subfigure}
		\hspace{-1em}
		\begin{subfigure}[b]{0.26\textwidth}
			\includegraphics[width=\textwidth]{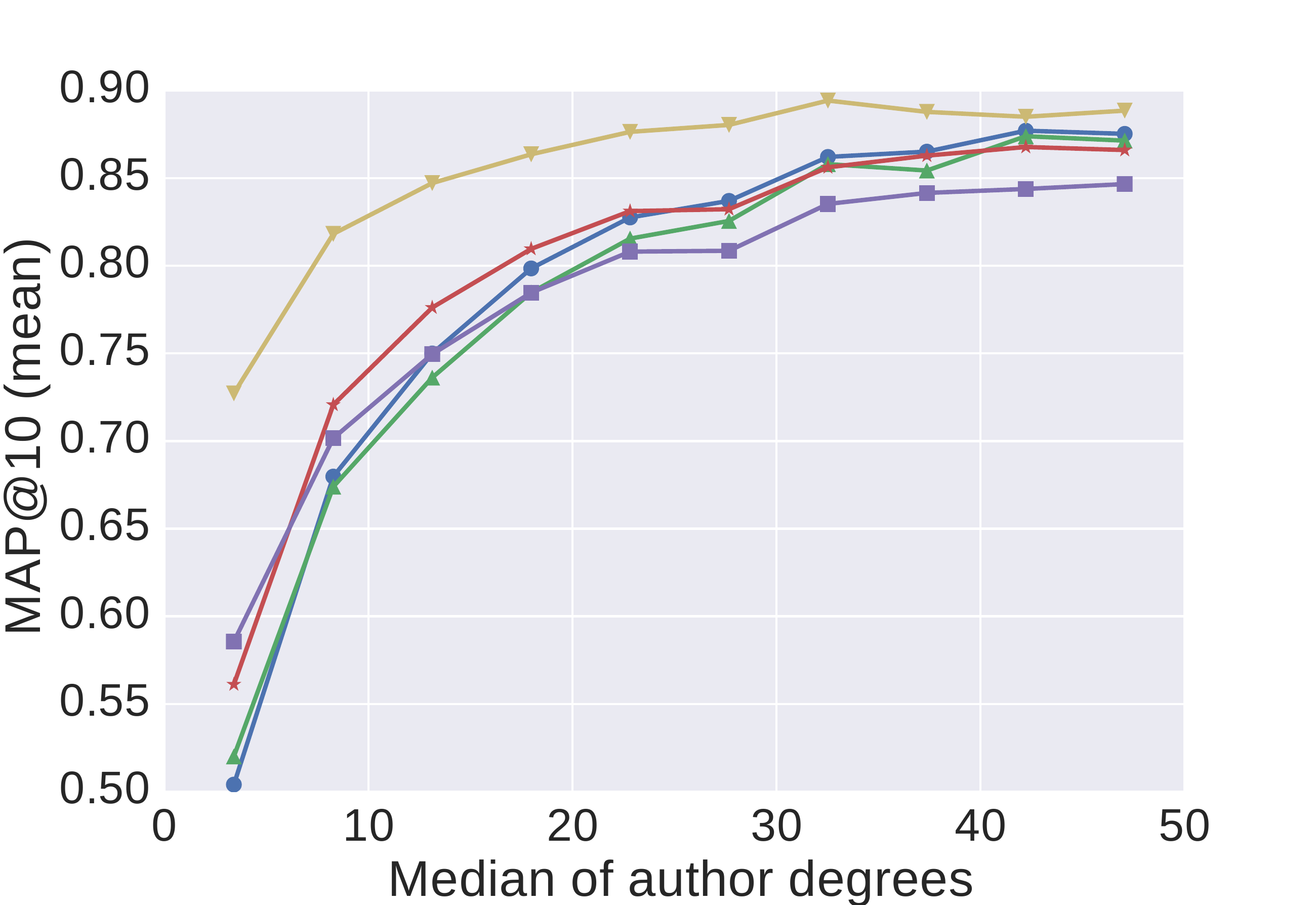}
			\caption{MAP@10}
		\end{subfigure}
		\hspace{-1em}
		\begin{subfigure}[b]{0.26\textwidth}
			\includegraphics[width=\textwidth]{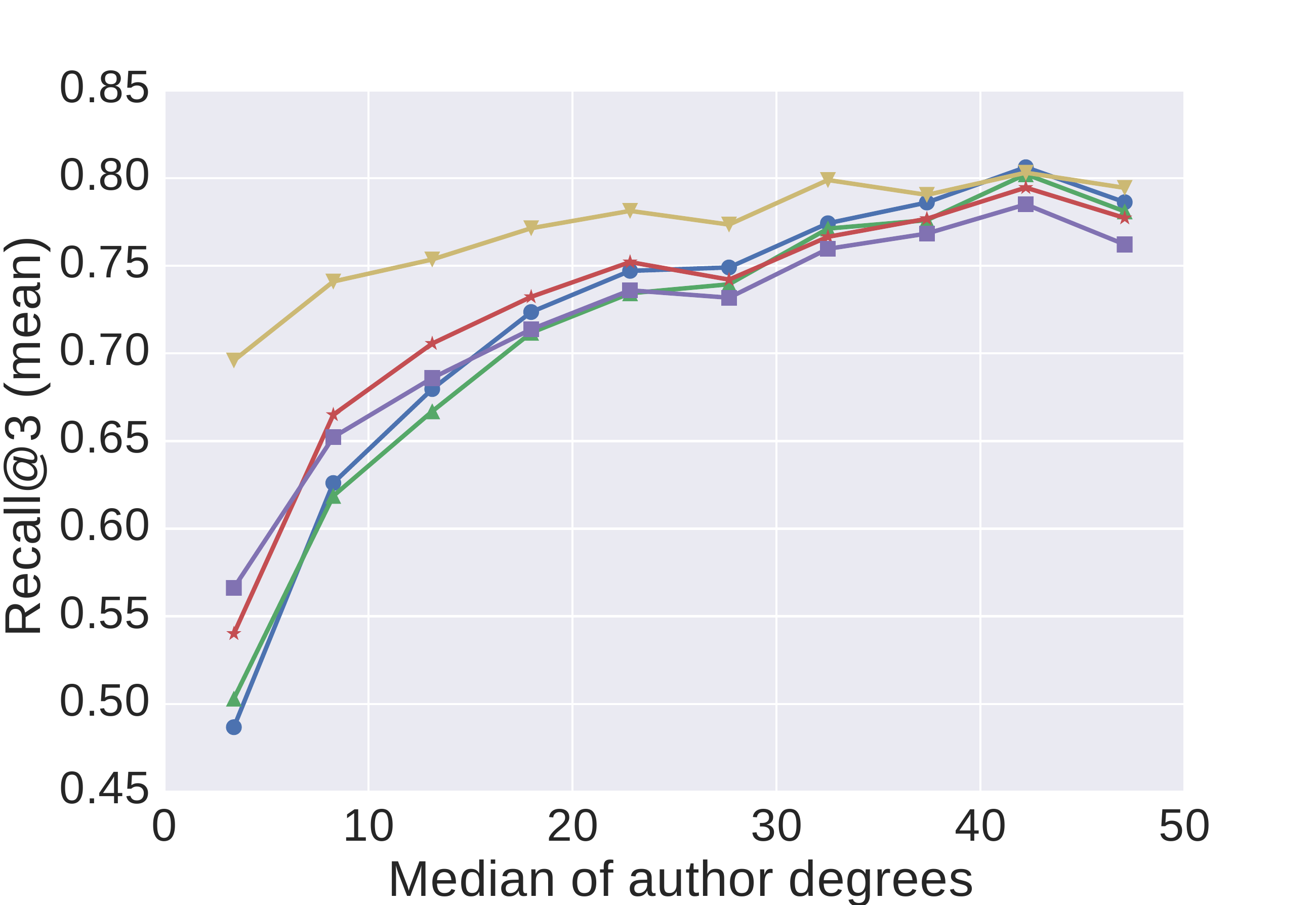}
			\caption{Recall@3}
		\end{subfigure}
		\hspace{-1em}
		\begin{subfigure}[b]{0.26\textwidth}
			\includegraphics[width=\textwidth]{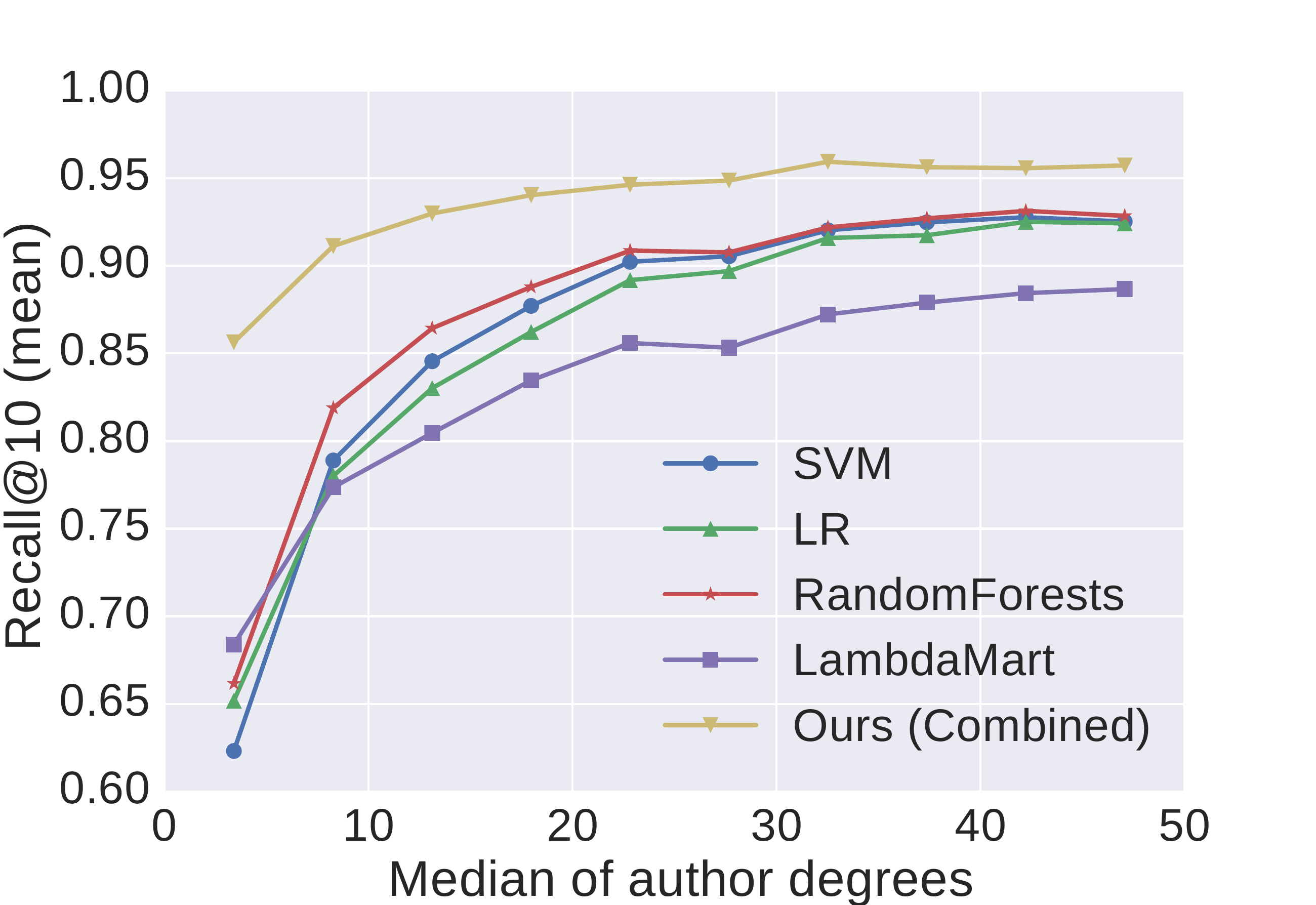}
			\caption{Recall@10}
		\end{subfigure}
	}
	\caption{Ranking results for author nodes of different degrees.}\label{fig:author_deg}
\end{figure*}

Figure \ref{fig:path_select_a} shows the results of single path performance, i.e., the performance when only a single meta-path is used in network-general embedding. Each dot in the plot indicates the performance of author prediction task for the validation dataset. The horizontal line indicates the performance of task-specific only embedding model. Note that paths are sorted according to their performance, and only paths that can help improve the author identification task are shown in the figure. 

Figure \ref{fig:path_select_b} shows the results of additive path selection, which demonstrate the performance of the combined model when meta-paths are added gradually. Each bar in the graph shows performance of the joint model based on specific additive selection of paths. Each single path is added to the network-general embedding sequentially according to their rank in the single path performance experiments. For example, the third bar with label ``+P1A" includes three paths: A-P-P, A-P-W, and P-A.

We observe the author identification performance grow first during the first several additive selection of paths, and then it starts to decrease as we add more paths. This suggests that first several paths are most relevant and helpful, and the latter ones can be less relevant, noisy, and thus they are harmful to use in network-general embedding. It also verifies our hypothesis that heterogeneous network embedding based on different meta paths will lead to different embeddings. Finally we select the first three paths A-P-P, A-P-W, and P-A in joint learning of the proposed model.

To further investigate the impact of using different meta paths on learning embeddings for the prediction task, we consider several types of paths: (1) the original length-1 network paths presented by network schema in Figure \ref{fig:net_schema}, (2) the augmented paths by combining all length-1 and length-2 paths, and (3) the selected paths by our procedure.

Table \ref{tab:path_comp} shows the results of different embedding models trained based on pre-given meta paths. We observe that by adding all length-2 paths, the results actually become worse, which might be due to the irrelevant or noisy paths. However, this does not mean that consider augmented paths are unnecessary. Using the greedy selected paths (A-P-P, A-P-W, and P-A) from both length-1 and length-2 paths, the performance of all models can be improved, which again demonstrate the path selection can play an important role in learning task-related embeddings.

\subsection{Performance Comparison with Baselines}

Table \ref{tab:result_tradition} shows the performance comparison between baselines and the proposed method. For both pre-train and network-general model, they do not have access to the task-specific path selection, so original length-1 network paths are used.

Our method significantly outperforms all baselines, including both supervised feature-based baselines and variants of embedding methods. To our surprise, the task-specific embedding model performs quite badly without pre-trained embedding vectors, significantly lower than other methods. We conjecture this is due to overfitting, and can be largely alleviated by pre-training or joint learning with unsupervised network-general embedding. 

To further examine the superior performance of our method compared with traditional methods, we group the papers by its medium author degrees\footnote{The author degree is calculated based on the number of papers he/she has published in training data.}, and report the results on each groups. Figure \ref{fig:author_deg} shows that our method outperforms baseline methods in almost all groups of papers, but most significantly in those papers that have less frequent authors. This suggests that our method can better understand authors with fewer links. For traditional feature based methods, it is very difficult to extract useful information/feature for them, but our model can still utilize propagation between authors and learn useful embeddings for them.

\begin{figure}[]
\begin{center}
\epsfig{file=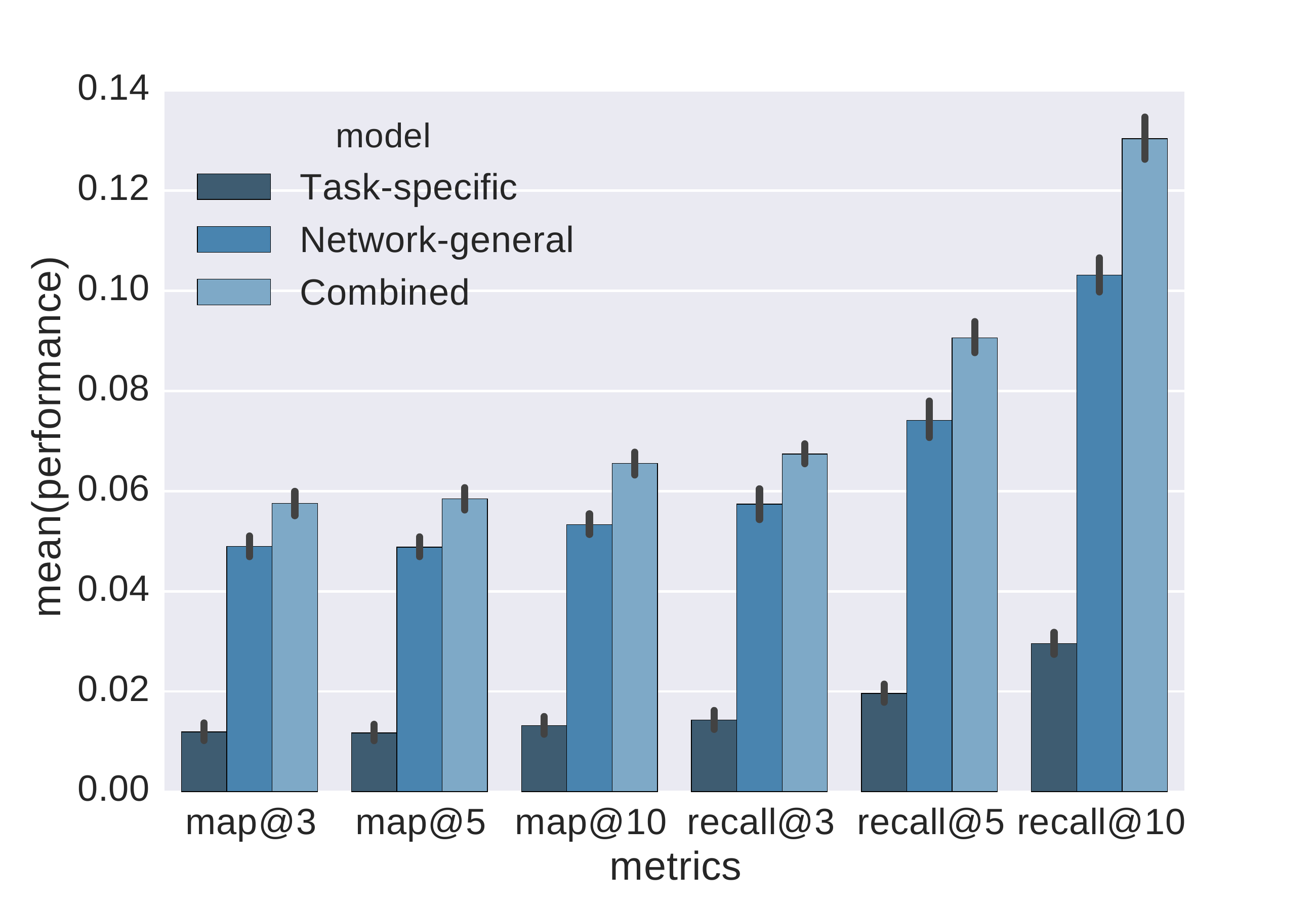, width=0.4\textwidth}
\end{center}
\caption{\label{fig:whole_candidate_set} Performance comparison on whole million authors candidate set.}
\end{figure}

\textbf{Whole author candidate set.} To test in real-world author prediction setting, we also conduct evaluation on the whole candidate set including a million of authors for variants of embedding methods. We only compare embedding methods as supervised feature based methods cannot scale up to whole candidate set. The results are shown in Figure \ref{fig:whole_candidate_set}. Due to the use of large candidate set, and thus longer evaluation time, we randomly sample 1000 test papers for a single experiment, and results are averaged over 10 experiments. We observe that, among variants of embedding methods, the combined method consistently outperforms other two variants. 

\begin{table}[!t]
	\small
	\centering
	\caption{Comparison of performance under different network paths (each entry is MAP@3 / Recall@3).}
	\label{tab:path_comp}
	\begin{tabular}{cccc}
		\Xhline{2\arrayrulewidth}
		{} &     Network-general &         Pre-train + Task &         Combined \\ \Xhline{2\arrayrulewidth}
		length-1       &  0.7563 / 0.7105 &  0.7722 / 0.7234 &   0.759 / 0.7133 \\
		length-(1+2)       &  0.7225 / 0.6847 &  0.7489 / 0.7082 &  0.7385 / 0.6973 \\
		Selected &  \textbf{0.7898} / \textbf{0.7379} &  \textbf{0.7914} / \textbf{0.7413} &  \textbf{0.8113} / \textbf{0.7548} \\
		\Xhline{2\arrayrulewidth}
	\end{tabular}
\end{table}

\begin{table}[t]
	\small
	\centering
	\caption{Author identification performance comparison.}
	\label{tab:result_tradition}
	\begin{tabular}{ccccccc}
		\Xhline{2\arrayrulewidth}
		Models         & MAP@3              & MAP@10          & Recall@3              & Recall@10       \\ \Xhline{2\arrayrulewidth}
		LR         &  0.7289 &  0.7321 &    0.6721 &     0.8209 \\
		SVM        &  0.7332 &  0.7365 &    0.6748 &    0.8267 \\
		RF         &  0.7509 &  0.7543 &    0.6921 &    0.8381 \\
		LambdaMart &  0.7511 &  0.7420 &    0.6869 &    0.8026 \\
		\hline
		Task-specific        &  0.6876 &  0.7088 &    0.6523 &    0.8298 \\
		Pre-train+Task.   &  0.7722 &  0.7962 &    0.7234 &    0.9014 \\
		Network-general   &  0.7563 &  0.7817 &    0.7105 &   0.8903 \\
		Combined &  \textbf{0.8113} &  \textbf{0.8309} &    \textbf{0.7548} &    \textbf{0.9215} \\
		\Xhline{2\arrayrulewidth}
	\end{tabular}
\end{table}

\subsection{Case Studies}

We show two types of case studies to demonstrate the performance differences between our proposed method and variants of embedding methods. The first type of case study shows the ranking of authors given some terms, which is used to see if the learned embedding nodes make sense. And the second type of case study shows the ranking of authors given information of anonymized paper, which is our original task.

Table \ref{tab:case_term} shows the ranking of authors given the term ``variational inference". We find from the results, the returned authors of combined methods are most reasonable (i.e., most likely to be the authors of the queried keyword), followed by general network embedding. And the task-specific embedding model itself sometimes give less reasonable results.

Table \ref{tab:case_paper} shows the ranked authors of some selected papers. Since the information provided for a paper is quite limited (keywords and limited references), and the number of whole candidate author set is more than one million, many of the true authors may not be presented in the top list. However, our combined method can predict true authors more accurately than other methods. Also, we find that most of the top authors in the returned list are related to the paper's topic and true authors, so it is sensible to consider them as potential authors of the paper.

\begin{table}[!t]
\small
\centering
\caption{Top ranked authors by models for queried keyword ``variational inference"}
\label{tab:case_term}
\begin{tabular}{p{2.6cm}p{2.6cm}p{2.6cm}}
\Xhline{2\arrayrulewidth}
Task-specific & Network-general & Combined \\
\Xhline{2\arrayrulewidth}
       Chong Wang &           Yee Whye Teh &     Michael I. Jordan \\
        Qiang Liu &  Mohammad E. Khan &          Yee Whye Teh \\
        Sheng Gao &         Edward Challis &     Zoubin Ghahramani \\
          Song Li &   Ruslan Salakhutdinov &  John William Paisley \\
      Donglai Zhu &      Michael I. Jordan &         David M. Blei \\
 Neil D. Lawrence &      Zoubin Ghahramani &           Max Welling \\
 Sotirios Chatzis &        Matthias Seeger &    Alexander T. Ihler \\
            Si Wu &        David B. Dunson &          Eric P. Xing \\
        Huan Wang &             Dae Il Kim &   Ryan Prescott Adams \\
       Weimin Liu &   Pradeep D. Ravikumar &   Thomas L. Griffiths \\
\Xhline{2\arrayrulewidth}
\end{tabular}
\end{table}

\begin{table}[!t]
\small
\centering
\caption{Top ranked authors for queried paper}
\label{tab:case_paper}

\begin{subtable}{0.5\textwidth}
\centering
\caption{``Active learning for networked data based on non-progressive diffusion model"}
\begin{tabular}{p{1.6cm}p{1.7cm}p{2.0cm}p{1.7cm}}
\Xhline{2\arrayrulewidth}
Ground-truth & Task-specific & Network-general & Combined\\
\Xhline{2\arrayrulewidth}
   \textbf{Z. Yang} &         L. Yu &         J. Leskovec &  \textbf{J. Tang} \\
      \textbf{J. Tang} &  Y. Gao  &             A. Ahmed &           H. Liu \\
        \textbf{B. Xu} &      J. Wang &           L. Getoor &         Y. Guo \\
 \textbf{C. Xing} &        H. Liu &           S.-D. Lin &       X. Shi \\
              \textbf{} &        Y. Gao &  D. Chakrabarti &            W. Fan \\
              \textbf{} &     Z. Wang &         P. Melville &           B. Zhang \\
              \textbf{} &   Z. Zhang &      T. Eliassi-Rad &        S.-D. Lin \\
              \textbf{} &         J. Zhu &           G. Lebanon &       H. Zha \\
              \textbf{} &    Y. Ye &            Y. Sun &      L. H. Ungar \\
              \textbf{} &       R. Pan &         L. H. Ungar &           C. Wang \\
\Xhline{2\arrayrulewidth}
\end{tabular}
\end{subtable}
\begin{subtable}{0.5\textwidth}
\centering
\caption{``CatchSync: catching sync. behavior in large directed graphs"}
\begin{tabular}{p{1.6cm}p{1.7cm}p{2.0cm}p{1.7cm}}
\Xhline{2\arrayrulewidth}
Ground-truth & Task-specific & Network-general & Combined\\
\Xhline{2\arrayrulewidth}

         \textbf{M. Jiang} &                     H. Wang &                 L. Akoglu &  \textbf{C. Faloutsos} \\
           \textbf{P. Cui} &                H. Tong &             T. Eliassi-Rad &             A. Gionis \\
        \textbf{A. Beutel} &  \textbf{C. Faloutsos} &                      U. Kang &                 L. Akoglu \\
 \textbf{C. Faloutsos} &         D. Chakrabarti &                H. Tong &             J. Kleinberg \\
      \textbf{S. Yang} &                H. Yang &         D. Chakrabarti &                J. Leskovec \\
                   \textbf{} &             G. Konidaris &             A. Gionis &         D. Chakrabarti \\
                   \textbf{} &             I. Stanton &                   X. Yan &               A. X. Zheng \\
                   \textbf{} &                     C. Wang &  \textbf{C. Faloutsos} &             T. Eliassi-Rad \\
                   \textbf{} &                    Y. Yang &                J. Leskovec &                      U. Kang \\
                   \textbf{} &                  S. Kale &   C. Tsourakakis &                H. Tong \\
\Xhline{2\arrayrulewidth}
\end{tabular}
\end{subtable}
\end{table}

\subsection{Parameter Study and Efficiency Test}

\begin{figure}[]
	\centering
	\makebox[1.35\linewidth][c]{%
		\hspace{-8em}
	\begin{subfigure}[b]{0.26\textwidth}
		\includegraphics[width=\textwidth]{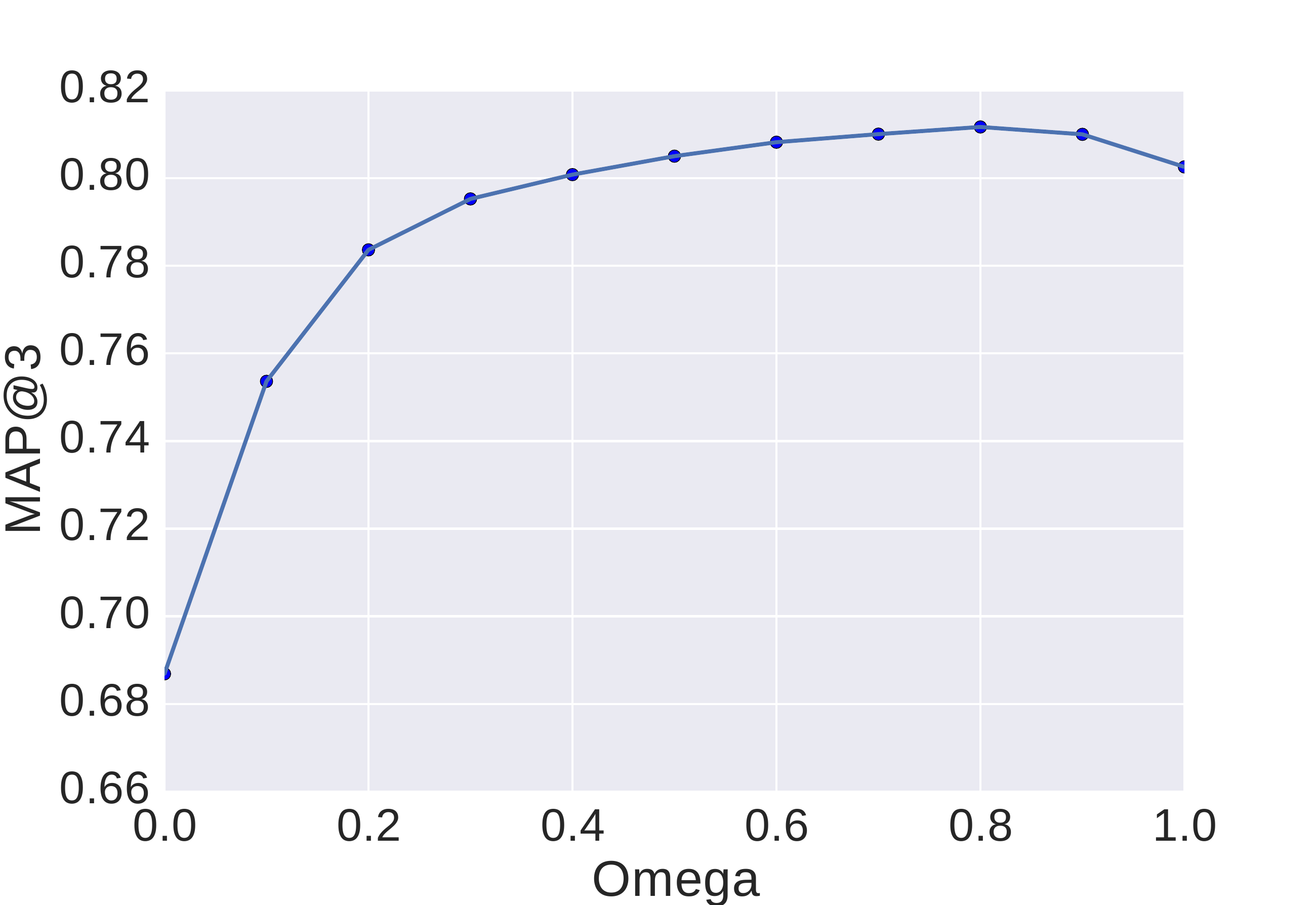}
		\caption{\label{fig:omega} }
	\end{subfigure}
	\hspace{-1em}
	\begin{subfigure}[b]{0.26\textwidth}
		\includegraphics[width=\textwidth]{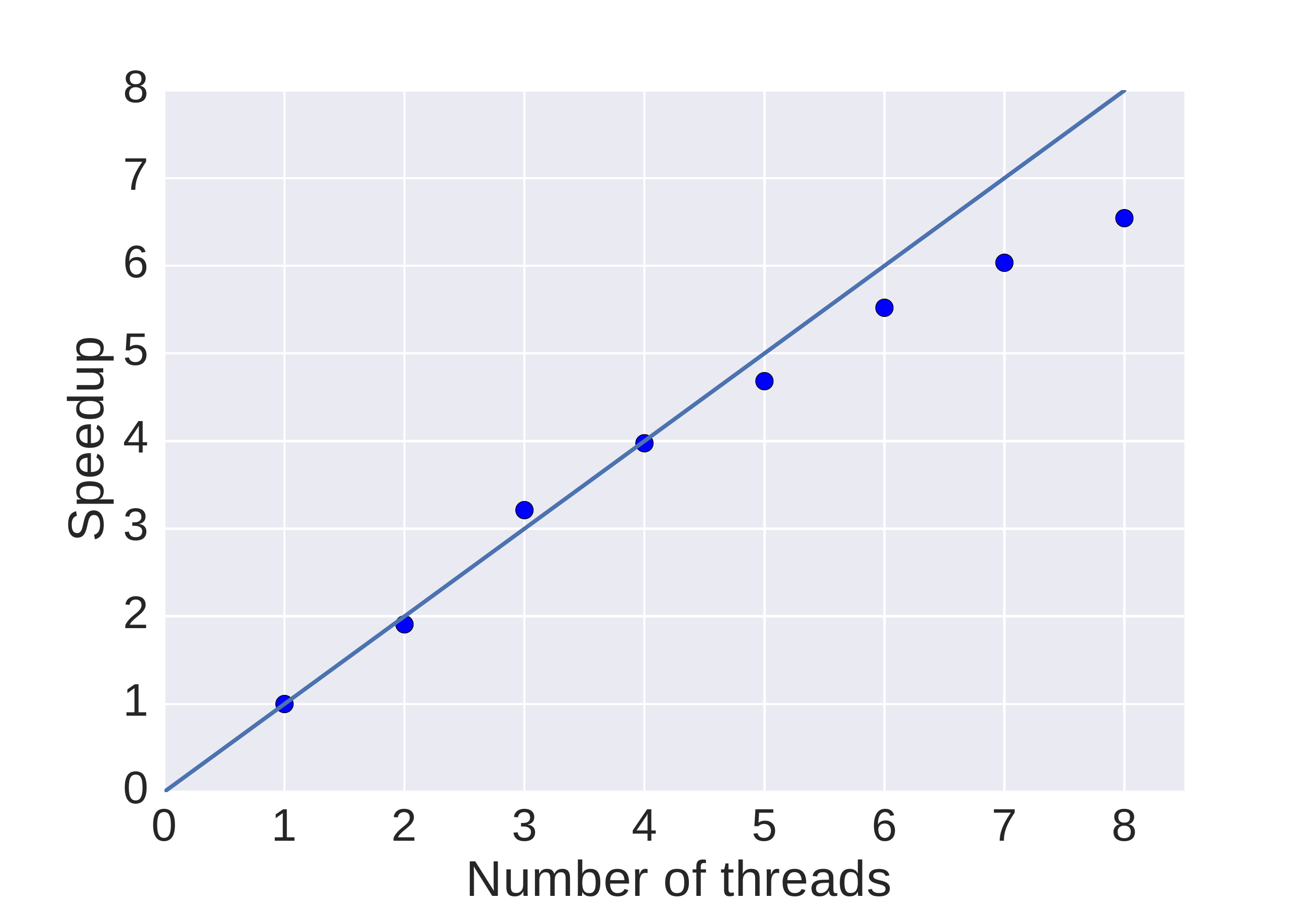}
		\caption{\label{fig:speed} }
	\end{subfigure}
	}
	\caption{(a) Choice of different combining factor between network-specific and network-general objectives. (b) Times of speed up versus the number of threads used.}
\end{figure}

We study the hyper-parameters $\omega$, which is the trade-off term for combing task-specific embedding and network-general embedding. The result is shown in Figure \ref{fig:omega}. As we can see that the best performance is obtained when we use $\omega=0.8$, at which both objectives are combined most appropriately.

Our model can be trained very efficiently with multi-core parallelization. All our experiments are conducted in a desktop with 4 core i7-5860k CPU and 64G memory. The experiments with embedding methods can be finished in about 10 minutes. To conduct a quantitatively experiment, we compare the times of training speed-up versus the number of threads used in Figure \ref{fig:speed}. It is almost linear speed-up for the first several number of threads, since our desktop CPU has only 4 cores (with hyper-threading), there are some overhead when the number of threads is more than 4.

\section{Discussion}

Although there is a severe lack of information about papers (e.g. the medium number of references per paper is 1, only keywords are used, and so on), our embedding based algorithm can still identify true authors with reasonable accuracy at top ranks, even with a million of candidate authors. We believe the model can be further improved by utilizing more complete information, and incorporating with more advanced text understanding techniques. For now and near future, a human expert can still be much more accurate at identifying the authors of a paper that he/she may be very familiar with, but algorithms may do a much better job when a paper is in some less familiar domains.

An interesting observation from both Figure \ref{fig:author_deg} and Table \ref{tab:case_paper} is that, authors with higher number of past publications are easier for the algorithm to predict, while the authors with few publication records are substantially harder.  This suggests that highly-visible authors may be easier to detect, while relatively junior researchers are harder to be identified. From this perspective, we think the double-blind review system is still helpful and in someway protects junior researchers.

\section{Related Work}

Many work has been devoted to mining heterogeneous networks in the past few years \cite{sun2012mining,sun2011pathsim,sun2013pathselclus,shi2015semantic}. To study such networks with multiple types of nodes and/or links, meta paths are proposed and studied \cite{sun2012mining,sun2011pathsim,sun2013pathselclus,shi2015semantic}. Many existing work on mining heterogeneous networks rely on feature engineering \cite{sun2011co,sun2012will}, while we adopt embedding methods for automatic feature learning.

Network embedding also attracts lots of attentions in recent years \cite{perozzi2014deepwalk,tang2015line,tang2015pte,chang2015heterogeneous,chen2016entity}. Many of these methods are technically inspired by word embedding \cite{mikolov2013distributed,mikolov2013efficient}. Different from traditional graph embedding methods \cite{yan2007graph}, such as multi-dimensional scaling \cite{cox2000multidimensional}, IsoMap \cite{tenenbaum2000global}, LLE \cite{roweis2000nonlinear}, Laplacian Eigenmap \cite{belkin2001laplacian}, the network embeddings are more scalable and shown better performance \cite{perozzi2014deepwalk,tang2015line}. Some existing network embedding methods are based on homogeneous network \cite{perozzi2014deepwalk,tang2015line}, while others are based on heterogeneous networks \cite{tang2015pte,chang2015heterogeneous}. Our work extends existing embedding methods by leveraging meta paths in heterogeneous networks, and use supervised task to guide the selection of meta paths.

The problem of author identification has been briefly studied before \cite{hill2003myth}. And we also notice KDD Cup 2013 has similar author identification/disambiguation problem \cite{li2013combination,li2013feature,zhong2013contextual,efimov2013kdd,zhao2013scorecard}, where participants are asked to predict which paper is truly written by some author. However, different from the KDD Cup, our setting is different from them in the sense that (1) existing authors are unknown in our double-blind setting, and (2) we consider the reference of the paper, which is one of the most important sources of information. Similar problems in social and information networks are also studied, such as collaboration prediction \cite{sun2011co,sun2012will}. The major difference between those work and ours is the methodology, their methods are mostly based on heavy feature engineering, while ours adopt automatic feature learning.

\section{Conclusion and Future Work}

In this paper, we study the problem of author identification under double-blind review setting, which is posed as author ranking problem under heterogeneous networks. To (1) embed network under the guidance of author identification task, and (2) better exploit heterogeneous networks with multiple types of nodes and links, we propose a task-guided and path-augmented heterogeneous network embedding model. In our model, nodes are first embedded as vectors in latent feature space. Embeddings are then shared and jointly trained by both task-specific and network-general objectives. We extend the existing unsupervised network embedding to incorporate meta paths in heterogeneous networks, and select paths according to the author identification task. The guidance is provided for learning network embedding, both explicitly in a joint objective and implicitly in path selection. Our experiments demonstrate the usefulness of meta paths in heterogeneous network embedding, and show that by combining both tasks, our model can obtain significantly better accuracy at identifying the true authors comparing to existing methods.

Some potential future work includes (1) author set prediction, where the interactions between authors will be considered in the prediction task, and (2) deeper analysis on text, given the full text of papers is given.


\section*{Acknowledgement}

We would like to thank anonymous reviewers for helpful suggestions. This work is partially supported by NSF CAREER \#1453800.

\bibliographystyle{abbrv}
\bibliography{main}

\appendix

\section{Feature Engineering for traditional supervised models}

For the traditional supervised models, we consider both author features and paper-author paired features for ranking authors given a paper. What follows we first show the author features we utilized.

\begin{itemize}
	\item Total number of papers
	\item Number of distinct venues
	\item Number of distinct years
\end{itemize}

There are four types of paper-author paired features being utilized, as shown below. 

\textbf{Paper references related}
\begin{itemize}
	\item Number of references being cited by the author before
	\item Ratio of references being cited by the author before
	\item Number of author's citations in the references
	\item Ratio of author's citations in the references
	\item Number of references written by the author
	\item Ratio of references written by the author
	\item Ratio of author's papers in the references
\end{itemize}

\textbf{Paper words related}
\begin{itemize}
	\item Number of shared word
	\item Number of unique shared word
	\item Ratio of shared words
	\item Ratio of unique shared words
\end{itemize}

\textbf{Paper venue related}
\begin{itemize}
	\item Whether the author attend the venue before
	\item Number of times the author attend the venue before
	\item Ratio of times the author attend the venue before
\end{itemize}

\textbf{Paper year related}
\begin{itemize}
	\item Number of papers author published in the last 3 years
	\item Ratio of papers author published in the last 3 years
\end{itemize}



\section{Derivation of Task-specific Embedding for Author Identification}

The gradients of the parameters in Task-specific Embedding model are calculated as follows.
\begin{equation}
\label{eq:sup_grad0}
\begin{gathered}
\nabla_{u_n} = \frac{w_t}{|X^{(t)}_p|}\bigg(\nabla_{f(p, a)}  u_{a} + \nabla_{f(p,a')} u_{a'}\bigg)
\\
\nabla_{u_a} = \nabla_{f(p, a)} \sum_t w_t \sum_{n\in X^{(t)}_p} u_n / |X^{(t)}_p|
\\
\nabla_{u_{a'}} = \nabla_{f(p, a')} \sum_t w_t \sum_{n\in X^{(t)}_p} u_n / |X^{(t)}_p|
\end{gathered}
\end{equation}
\begin{equation}
\label{eq:sup_grad1}
\begin{split}
\nabla_{w_t} =\bigg(\nabla_{f(p,a)} u_a + \nabla_{f(p,a')} u_{a'}\bigg)^T \bigg( \sum_{n\in X^{(t)}_p} u_n / |X^{(t)}_p| \bigg)
\end{split}
\end{equation}
where
$$
\nabla_{f(p,a)} = \delta(f(p,a') - f(p,a) + \epsilon)
$$$$
\nabla_{f(p,a')} = -\delta(f(p,a') - f(p,a) + \epsilon)
$$
where $\delta(x)$ is an indicator function, which is set one if and only if $x$ is greater than 0.

The learning algorithm is illustrated in Algorithm \ref{alg:sup}.

\begin{algorithm}[]
	\caption{Task-specific embedding for author identification}
	\label{alg:sup}
	\renewcommand{\algorithmicrequire}{\textbf{Input:}}
	\renewcommand\algorithmicensure {\textbf{Output:}}
	\begin{algorithmic}[1]
		\REQUIRE paper information $X$, and true author set $A$
		\ENSURE parameters $U$, $w$
		\WHILE{not converged}
		\FOR{each thread}
		\STATE sample a triple $(p, a, a')$
		\STATE update $U, w$ according to Eq. \ref{eq:sup_grad0}, \ref{eq:sup_grad1}
		\ENDFOR
		\ENDWHILE
	\end{algorithmic}
\end{algorithm}

\section{Derivation of Path-augmented General Heterogeneous Network Embedding}

The gradient of the parameters in Path-augmented General Heterogeneous Network Embedding model can be calculated as follows.
\begin{equation}
\label{eq:unsup_grad0}
\begin{gathered}
\nabla_{u_i} = (1-\sigma(u_i^T u_j + b_r)) u_j - \sigma(u_i^T u_{j'} + b_r) u_{j'} \\
\nabla_{u_j} = (1-\sigma(u_i^T u_j + b_r)) u_i \\
\nabla_{u_{j'}} = - \sigma(u_i^T u_{j'} + b_r) u_i \\
\end{gathered}
\end{equation}
\begin{equation}
\label{eq:unsup_grad1}
\nabla_{b_r} = (1-\sigma(u_i^T u_j + b_r)) - \sum_{j'} \sigma(u_i^T u_{j'} + b_r)
\end{equation}
The learning algorithm of the model is summarized in the Algorithm \ref{alg:unsup}.

\begin{algorithm}[]
	\caption{Path-augmented general heterogeneous network embedding}
	\label{alg:unsup}
	\renewcommand{\algorithmicrequire}{\textbf{Input:}}
	\renewcommand\algorithmicensure {\textbf{Output:}}
	\begin{algorithmic}[1]
		\REQUIRE paths adjacency matrices $\{M\}$ derived from the heterogeneous network $G$.
		\ENSURE parameters $U, b$
		\WHILE{not converged}
		\FOR{each thread}
		\STATE sample a triple $(r, i, j)$
		\STATE sample negative nodes $\{j'\}$
		\STATE update $U, b$ according to Eq. \ref{eq:unsup_grad0}, \ref{eq:unsup_grad1}
		\ENDFOR
		\ENDWHILE
	\end{algorithmic}
\end{algorithm}


\end{document}